\documentclass[letterpaper, 10 pt, conference]{ieeeconf}  

\IEEEoverridecommandlockouts                              

\usepackage{algorithm}
\usepackage{algpseudocode}
\usepackage{array} 
\usepackage{graphicx}
\usepackage{amsmath} 
\usepackage{comment}
\usepackage{booktabs}       
\usepackage{multirow}
\usepackage{makecell}
\usepackage{adjustbox}
\newcommand{\best}[1]{\textbf{#1}}
\usepackage{xcolor}
\usepackage{cite}
\usepackage{kotex}

\usepackage{amsfonts}
\usepackage{amssymb}
\usepackage{txfonts}

\usepackage{booktabs}
\usepackage{nicematrix}

\newcolumntype{C}{>{\centering\arraybackslash}p{4.5em}}

\newcommand{\method}{\textsc{SARAM}}

\title{\Large \bf{
SARAM: Sparse Adapter Allocation with Reuse and Merging\\ for Lifelong Robot Learning}
}

\author{SeongJun Jeong$^1$, Chungwoo Lee$^1$, Gi-Cheon Kang$^{2*}$, and Byoung-Tak Zhang$^{1*}$%
\thanks{$^*$Corresponding Authors}%
\thanks{$^{1}$Seoul National University}%
\thanks{$^{2}$Ajou University}%
\thanks{
This work was partly supported by grants from the IITP (RS-2021-II211343-GSAI/10\%, RS-2022-II220951-LBA/10\%, RS-2022-II220953-PICA/10\%, RS-2026-25553157-MIACC/10\%), the NRF (RS-2024-00353991-SPARC/15\%, RS-2023-00274280-HEI/15\%), the KEIT (RS-2025-25453780/15\%), and the KIAT (RS-2025-25460896/15\%), funded by the Korean government.
}}

\begin{document}

\maketitle
\thispagestyle{empty}
\pagestyle{empty}


\begin{abstract}
Lifelong robot learning requires acquiring new tasks sequentially without forgetting earlier ones.
With large pre-trained vision-language-action (VLA) backbones, adapter-based methods effectively mitigate this challenge.
However, their task-specific adapters and selection modules accumulate with the task stream, creating a scalability challenge in on-disk storage and deployment cost under fixed onboard resource budgets.
We propose \method{}, a lifelong learning framework that limits storage growth while keeping GPU memory and per-step inference cost independent of the task stream length. 
For each new task, \method{} reuses frozen adapters from previous tasks and introduces new adapters only at sparsely selected layers, with a duplication penalty encouraging them to capture information not already represented by the adapter library. 
At deployment, a compact signature derived from backbone activation statistics identifies the task once per episode, and the corresponding task update is merged into the backbone, leaving only the backbone in the control loop.
Across four LIBERO suites, \method{} matches or exceeds adapter-based baselines while adding up to $4.5\times$ less on-disk state per task. 
GPU memory usage and control frequency remain comparable to those of the backbone as the task stream grows. 
Experiments on a real UR5 arm with a different backbone further demonstrate retention throughout a sequential task stream.
\end{abstract}

\section{INTRODUCTION}
\label{sec:intro}
Robots deployed in real environments need to learn new tasks continually~\cite{liu2023libero}, yet onboard hardware imposes fixed budgets on storage, memory, and compute.
Learning tasks sequentially causes catastrophic forgetting~\cite{goodfellow2013empirical,wang2024comprehensive}, where acquiring a new task degrades performance on previously learned ones.
Existing approaches address this through replay, regularization, or architectural isolation, which require storing past demonstrations, provide weaker protection against forgetting, or leave progressively less free capacity for new tasks, respectively~\cite{chaudhry2019tiny,saha2021gradient,kirkpatrick2017overcoming,mallya2018packnet}.
With large pre-trained vision-language-action (VLA) backbones~\cite{kim2024openvla,black2024pi,brohan2023rt}, recent methods~\cite{liu2024tail,romer2026clare,lei2025dynamic} increasingly adopt lightweight adapters such as LoRA~\cite{hu2021lora}.
The backbone remains frozen, while each task learns its own adapters, which are then frozen so that later training cannot overwrite them. 
By freezing the backbone and task-specific adapters, these methods effectively preserve previously learned tasks.

However, task-specific adapters must be selected at deployment, and the oracle task ID is typically unavailable.
Instead, the robot receives the current observation and language instruction, from which the appropriate task-specific adapter must be identified.
Recent methods~\cite{romer2026clare,lei2025dynamic} dynamically select task-specific adapters from the current input during execution.
By design, the adapters and selection modules therefore remain active in the control loop, increasing GPU memory and per-step computation as tasks accumulate.

\begin{figure}[t]
\centering
\includegraphics[width=\linewidth]{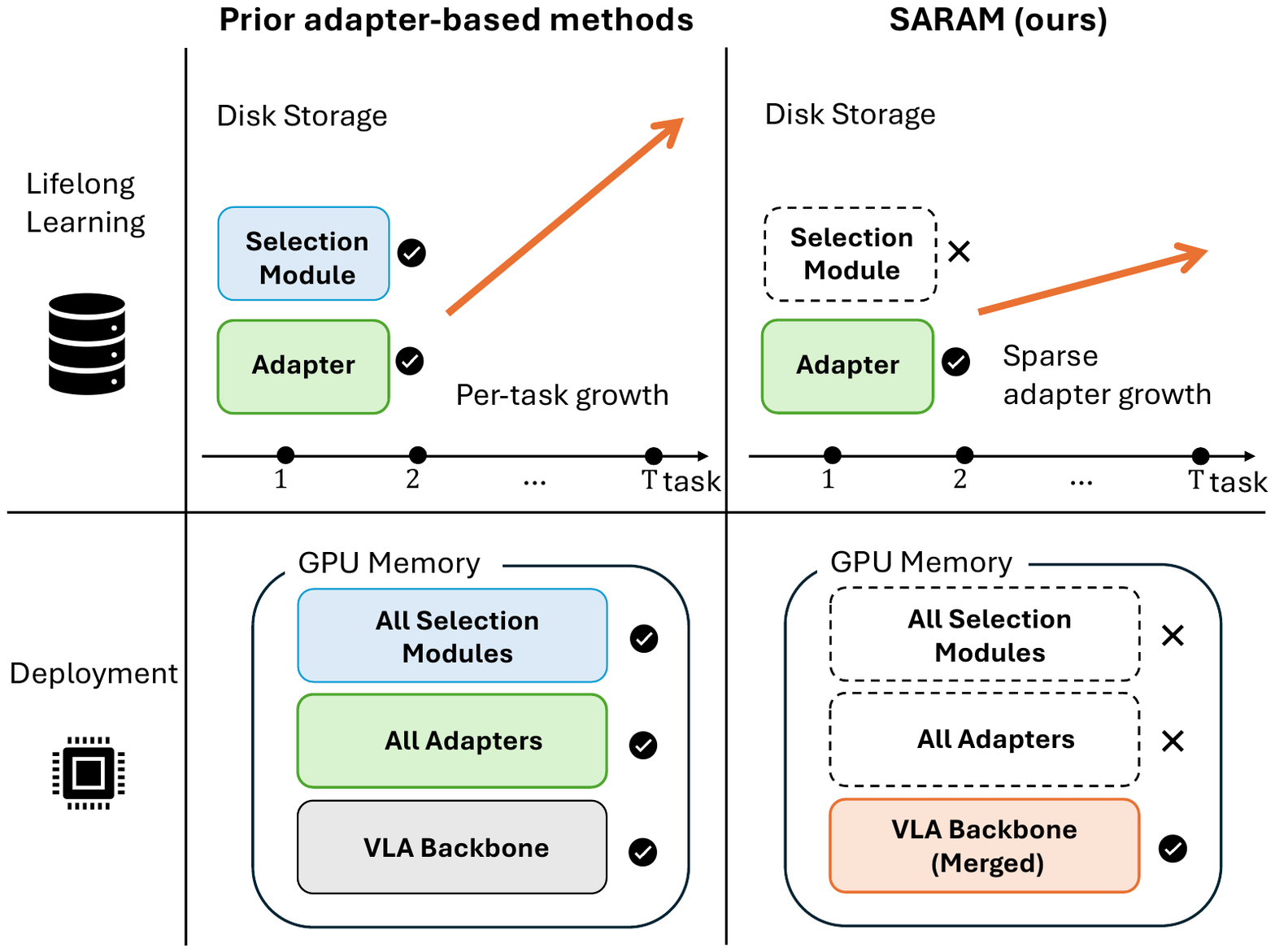}
\caption{\textbf{Two growth axes of adapter-based lifelong robot learning.}
Prior methods accumulate adapters and selection modules on disk and keep them available on the GPU during control.
\method{} adds new adapters sparsely and deploys a single merged backbone, adding up to $4.5\times$ less on-disk state per task in our experiments.}
\label{fig:motivation}
\end{figure}

As illustrated in Fig.~\ref{fig:motivation}, this accumulation appears along two axes.
During training, task-specific adapters, selection modules, and associated state accumulate on disk as new tasks are learned.
At deployment, the adapters and selection modules must remain available on the GPU and are used throughout the control loop.
Both axes therefore grow with the task stream, conflicting with the fixed resource budgets of onboard robot hardware.
\emph{Can a VLA policy learn new tasks continually while keeping its deployment cost independent of the task stream and limiting task-dependent storage growth?}

To address this, we propose \method{}, a lifelong robot learning method that addresses the two axes at their respective stages: storage as tasks accumulate, and deployment cost at execution.
During learning, each new task first reuses a linear combination of frozen LoRA adapters from previous tasks, then adds new adapters only at layers selected under a sparsity penalty.
A duplication penalty discourages the new adapters from reproducing what the reused library already expresses.
The newly allocated adapters and combination coefficients are frozen after training, so later tasks cannot overwrite previously learned parameters. Because every per-task update stays linear in the weights, it can later be folded into the backbone rather than evaluated as a separate module.

\method{} then removes task selection from the control loop. 
After each task is learned, the demonstrations are passed through the adapter-free backbone and its layer-wise activation statistics are summarized into a compact task signature.
These frozen-backbone statistics provide task-dependent signatures without requiring a learned selection module.
At deployment, the initial observation and instruction are summarized in the same way, matched against the stored signatures, and the selected update is merged into the backbone before execution. 
The episode therefore runs on a single backbone, with no task-specific adapter or selection module inside the control loop, keeping GPU memory and per-step inference cost at the backbone's level regardless of the number of learned tasks.

We evaluate \method{} on four LIBERO suites against adapter-based methods~\cite{lei2025dynamic,romer2026clare} and standard continual learning baselines.
\method{} attains the highest final success rate and AUC on all four suites while requiring up to $4.5\times$ less on-disk state per task than adapter-based baselines.
At deployment, \method{} remains at the backbone's level in both GPU memory and control frequency: after ten tasks it runs at 15.2\,Hz with 748\,MB of GPU memory, identical to the backbone alone.
By contrast, CLARE and DMPEL grow to 922\,MB and 829\,MB of GPU memory and slow to 10.1\,Hz and 5.0\,Hz after ten tasks, respectively.
On a seven-task real-robot stream with a different VLA backbone, \method{} ends the stream at the same success rate it reached when each task was learned, showing that the method transfers to a physical robot and a different backbone architecture.

We summarize our contributions as follows:
\begin{itemize}
    \item We propose \method{}, an adapter-based lifelong robot learning method that limits per-task storage through adapter reuse, sparse allocation, and compact task signatures in place of learned selection modules.
    \item At deployment, task signatures move selection outside the control loop, and the corresponding task update is merged into the backbone to keep GPU memory and per-step inference cost independent of the number of learned tasks.
    \item On four LIBERO suites, \method{} attains the highest final success rate and AUC while requiring up to $4.5\times$ less on-disk state per task, and real-robot experiments with a different VLA backbone demonstrate applicability beyond simulation.
\end{itemize}

\section{RELATED WORK}
\label{sec:related}

\noindent \textbf{Lifelong Robot Learning.}
Lifelong robot learning~\cite{thrun1995lifelong,liu2023libero} studies how robots can learn new tasks sequentially while retaining those learned before.
A central challenge, shared with continual learning in general~\cite{wang2024comprehensive}, is catastrophic forgetting~\cite{kirkpatrick2017overcoming}, where learning new tasks degrades performance on earlier ones.
Existing approaches address this problem through replay, regularization, or architectural isolation.
Replay methods retain past demonstrations for later training~\cite{chaudhry2019tiny}, which incurs storage overhead and can be undesirable when past demonstrations cannot be retained.
Regularization methods such as EWC~\cite{kirkpatrick2017overcoming} avoid storing past data but often provide weaker protection against forgetting.
Architectural-isolation methods such as PackNet~\cite{mallya2018packnet} protect previous tasks by reserving separate capacity, but the available capacity becomes increasingly constrained as more tasks are learned.
In robot manipulation, LIBERO~\cite{liu2023libero} provides a standard benchmark for evaluating sequential task learning and retention.

\noindent \textbf{Adapter-Based Lifelong Robot Learning.}
With the emergence of large pretrained VLA policies~\cite{black2024pi,kim2024openvla,intelligence2025pi,brohan2022rt,brohan2023rt}, recent work has increasingly adopted lightweight adapters for task-specific adaptation while keeping the backbone frozen~\cite{liu2024tail,romer2026clare,lei2025dynamic}.
After each task is learned, its adapter is frozen, preventing later training from overwriting previously learned task parameters.
This provides strong retention, but requires maintaining additional task-specific parameters as the task stream grows.
When the oracle task ID is unavailable, CLARE~\cite{romer2026clare} and DMPEL~\cite{lei2025dynamic} dynamically select or combine task-specific modules from the current input using learned discriminators and a router, respectively.
These input-dependent mechanisms remain active during control, causing task-specific storage and deployment cost to grow with the task stream.

\noindent \textbf{Resource-Efficient Continual Learning.}
Recent work beyond robotics has explored reducing the parameter growth of adapter-based continual learning.
Some approaches compress task-specific adapters into a shared low-rank representation or merge multiple task updates into a shared model~\cite{kaushik2026shared,qiao2026merge,fukuda2025adapter}.
These methods reduce parameter growth, but typically update shared parameters that also represent previously learned tasks.
Their evaluation settings typically do not consider task selection from observations and instructions at deployment.
In contrast, \method{} keeps previously learned task parameters frozen, reuses past adapters through combination coefficients, and adds new adapters only at layers retained by sparse allocation.
It further replaces learned selection modules with task signatures and merges the corresponding task update into the VLA backbone before execution, so only the backbone runs during control.

\section{METHOD}
\label{sec:method}

\begin{figure*}[t]
    \centering
    \includegraphics[width=\textwidth]{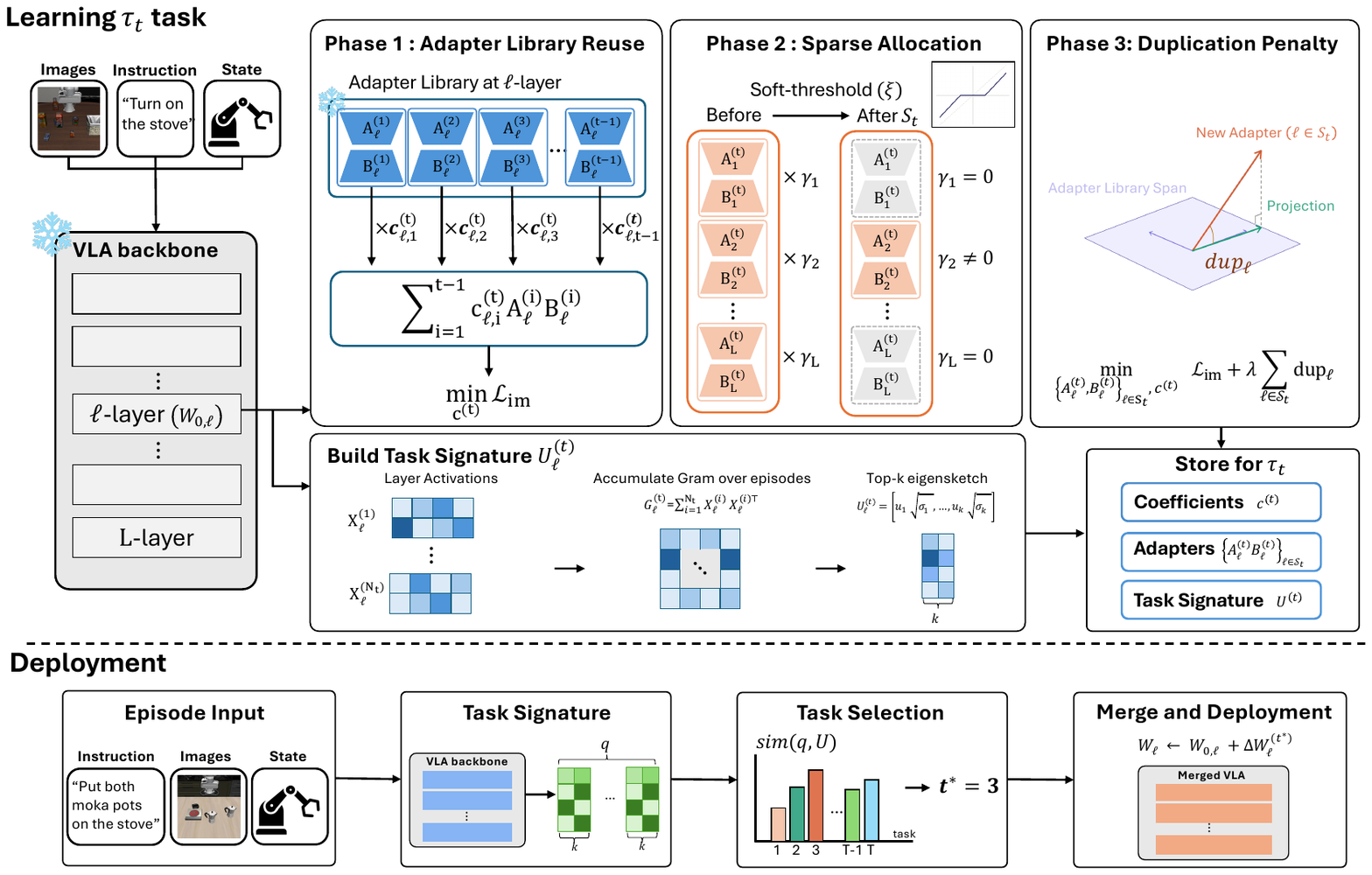}
    \caption{\textbf{Overview of \method{}.}
    Top: each task reuses the frozen library (Phase 1), allocates new adapters to a sparse subset of layers (Phase 2), trains them under a duplication penalty (Phase 3), and stores coefficients, adapters, and a task signature.
    Bottom: at deployment, one adapter-free pass selects the task and its update is merged, so only the backbone runs.}
    \label{fig:method}
\end{figure*}

\subsection{Problem Setting}
\label{sec:setting}
We follow the lifelong imitation learning setup of LIBERO~\cite{liu2023libero}.
A robot learns $T$ manipulation tasks $\{\tau_1,\dots,\tau_T\}$ sequentially with a single policy $\pi$ conditioned on observations and language instructions.
Each task $\tau_t$ comes with a language instruction $l_t$ and a demonstration dataset $\mathcal{D}_t=\{\mathcal{E}_i^{(t)}\}_{i=1}^{N_t}$, where each episode $\mathcal{E}_i^{(t)}$ contains a sequence of observation--action pairs.
Past datasets are not available unless the method stores them itself, at its own storage cost.
After each task is learned, the policy is evaluated on the current task and all tasks learned so far.
At deployment the task identity is not given, so the policy receives only the current observation and instruction.
The objective is therefore to acquire each new task while retaining performance on earlier ones.

\subsection{Overview}
\label{sec:overview}
\method{} addresses the two growth axes at their respective stages: disk storage during learning, and GPU cost at deployment. Each new task first reuses the frozen adapters of past tasks (Sec.~\ref{sec:prefit}).
A sparse adapter allocation then adds new capacity only at layers selected under a sparsity penalty (Sec.~\ref{sec:allocation}), and a duplication penalty keeps that capacity from re-expressing what the reused adapters already provide (Sec.~\ref{sec:training}). At deployment, every per-task update is linear in the weights, and the task is selected once per episode from frozen-backbone activation statistics (Sec.~\ref{sec:selection}). The selected update can therefore be added directly to the backbone weights, and the episode runs on a single backbone (Sec.~\ref{sec:merging}), holding GPU memory and per-step inference cost at the backbone's own level.

\subsection{Per-Task Updates as Linear Deltas}
\label{sec:delta}

\method{} adapts a pre-trained VLA backbone with frozen weights
$W_0=\{W_{0,\ell}\}_{\ell=1}^{L}$, where
$W_{0,\ell}\in\mathbb{R}^{d_\ell\times d'_\ell}$
and $\ell$ indexes the linear layers subject to adaptation.
After learning task $\tau_j$, its new adapters are kept as frozen rank-$r$ LoRA~\cite{hu2021lora} adapters, with
$A^{(j)}_\ell \in \mathbb{R}^{d_\ell \times r}$ and
$B^{(j)}_\ell \in \mathbb{R}^{r \times d'_\ell}$.
We refer to the adapters kept by $\tau_1,\dots,\tau_{t-1}$ as the library available to task $\tau_t$.

Let $\mathcal{S}_j$ denote the set of layers at which task $\tau_j$ received a new adapter, and let
$\mathcal{J}^{(t)}_\ell=\{j<t:\ell\in\mathcal{S}_j\}$
denote the previous tasks that have an adapter at layer $\ell$.
The update of task $\tau_t$ at layer $\ell$ combines the library adapters with a new adapter if one is allocated at that layer:
\begin{equation}
\Delta W^{(t)}_\ell
=
\sum_{j\in\mathcal{J}^{(t)}_\ell}
c^{(t)}_{\ell,j} A^{(j)}_\ell B^{(j)}_\ell
+
\mathbf{1}[\ell\in\mathcal{S}_t]
A^{(t)}_\ell B^{(t)}_\ell,
\label{eq:update}
\end{equation}
where $c^{(t)}_{\ell,j} \in \mathbb{R}$ is a combination coefficient and
$c^{(t)}=\{c^{(t)}_{\ell,j}:\ell\in[L],\,j\in\mathcal{J}^{(t)}_\ell\}$.

The two terms in Eq.~\eqref{eq:update} play different roles.
The first reuses frozen adapters from previous tasks through their combination coefficients, while the second introduces task-specific capacity through a new rank-$r$ adapter only when $\ell\in\mathcal{S}_t$.
Thus, each task update is decomposed into a component reused from the existing library and a component represented by newly allocated adapters.
As illustrated in Fig.~\ref{fig:method}, \method{} first exploits the former and allocates the latter only at selected layers.

For $t=1$, $\mathcal{J}^{(1)}_\ell$ is empty and the first term in Eq.~\eqref{eq:update} is zero.
The per-task parameters of $\tau_t$ are therefore
$\theta_t=\{c^{(t)},(A^{(t)}_\ell,B^{(t)}_\ell)_{\ell\in\mathcal{S}_t}\}$.
Because Eq.~\eqref{eq:update} defines a linear weight update with no intervening nonlinearity, $\Delta W^{(t)}_\ell$ can be added directly to $W_{0,\ell}$ rather than evaluated as a separate module.
Learning $\tau_t$ fills in this decomposition in three phases:
Phase~1 fits the combination coefficients $c^{(t)}$,
Phase~2 determines $\mathcal{S}_t$,
and Phase~3 trains the new adapters jointly with $c^{(t)}$.

\subsection{Phase 1: Adapter Library Reuse}
\label{sec:prefit}

The first phase fits the new task with the library alone, before any new capacity is added.
All new adapters are disabled, so that Eq.~\eqref{eq:update} reduces to a linear combination of past adapters, and only $c^{(t)}$ is trained under the backbone's imitation loss $\mathcal{L}_{\mathrm{im}}$, stopping when the loss plateaus.
The remaining error indicates what the library alone cannot express, and the later phases add new adapters to cover it.

\subsection{Phase 2: Sparse Adapter Allocation}
\label{sec:allocation}
The second phase determines which layers retain a new adapter.
A candidate adapter and a scalar gate $\gamma_\ell$ are attached to every layer, where the gate scales the candidate output and $\gamma_\ell=0$ removes the adapter from the update.
With the Phase~1 coefficients fixed, all candidates are trained jointly for a short allocation phase under the imitation loss with an $\ell_1$ penalty on the gates,
\begin{equation}
\min_{\{A^{(t)}_\ell, B^{(t)}_\ell, \gamma_\ell\}}
\mathcal{L}_{\mathrm{im}}
+ \xi \sum_{\ell} |\gamma_\ell|,
\label{eq:gate}
\end{equation}
where $\xi$ controls the strength of the sparsity penalty.
We optimize this objective by proximal gradient descent~\cite{beck2009fast,ding2023sparse}.
After a short warm-up without thresholding, each gradient step on $\mathcal{L}_{\mathrm{im}}$ is followed by the proximal operator of the $\ell_1$ term,
\begin{equation}
\gamma_\ell
\leftarrow
\mathrm{sign}(\gamma_\ell)
\max\!\left(|\gamma_\ell|-\eta\xi,\,0\right),
\label{eq:prox}
\end{equation}
where $\eta$ is the step size.
Weight decay is applied to $A^{(t)}_\ell$ and $B^{(t)}_\ell$ but not to $\gamma_\ell$, preventing a smaller gate from being compensated by larger LoRA factors.
We use the same gate initialization for every task, providing a common starting point for layer allocation while allowing task-specific imitation gradients to determine which gates survive.
The support $\mathcal{S}_t=\{\ell:\gamma_\ell\neq0\}$ is read off at the end of the allocation phase, and candidates outside $\mathcal{S}_t$ are discarded.

\subsection{Phase 3: Training with a Duplication Penalty}
\label{sec:training}

In the third phase, the surviving adapters and combination coefficients are trained jointly.
The objective adds to $\mathcal{L}_{\mathrm{im}}$ a duplication penalty that measures the overlap between a new adapter and the span of the library,
\begin{equation}
\begin{aligned}
&\min_{\{A^{(t)}_\ell,B^{(t)}_\ell\}_{\ell\in\mathcal{S}_t},\,c^{(t)}}
\mathcal{L}_{\mathrm{im}}
+ \lambda \frac{1}{|\mathcal{S}_t|}
\sum_{\ell\in\mathcal{S}_t}\mathrm{dup}_\ell, \\
&\qquad
\mathrm{dup}_\ell
=
\frac{b_\ell^\top M_\ell^{-1}b_\ell}
{\|A^{(t)}_\ell B^{(t)}_\ell\|_F^2},
\end{aligned}
\label{eq:dup}
\end{equation}
where
$M_\ell[i,j]
=
\langle A^{(i)}_\ell B^{(i)}_\ell,
A^{(j)}_\ell B^{(j)}_\ell\rangle$
contains the pairwise inner products among library adapters, and
$b_\ell[i]
=
\langle A^{(i)}_\ell B^{(i)}_\ell,
A^{(t)}_\ell B^{(t)}_\ell\rangle$
their inner products with the new adapter.
The numerator measures the component of the new adapter explained by the library span, and $\lambda$ controls how strongly such overlap is penalized.
This encourages reusable structure to be represented through $c^{(t)}$ while reserving new adapters for complementary directions.
For numerical stability, a small ridge proportional to the largest diagonal entry is added to $M_\ell$ before inversion.
Since the library is frozen, this inverse is computed once per task.
For the first task, or for a layer with no existing library adapter, the reuse term and duplication penalty are omitted.
After training, $\theta_t$ and the task signature described next are stored and kept frozen for the remainder of the stream.
Later tasks add only their own coefficients and adapters, leaving all previously learned task parameters unchanged.

\subsection{Task Selection from Task Signatures}
\label{sec:selection}

\method{} operates without the oracle task ID.
At deployment, the task is identified from the initial observation and instruction using activation statistics of the frozen backbone.
For each layer, we summarize its input activations through a Gram matrix, which captures task-dependent activation structure.
To build the signature of $\tau_t$, the initial observation and instruction of each demonstration are passed through the adapter-free backbone, and their token-level activations are accumulated as
\begin{equation}
\begin{aligned}
G_\ell^{(t)}
&=
\sum_{i=1}^{N_t}
X_\ell^{(i)}X_\ell^{(i)\top}
\in\mathbb{R}^{d_\ell\times d_\ell}, \\
U_\ell^{(t)}
&=
\big[
u_1\sqrt{\sigma_1},\dots,u_k\sqrt{\sigma_k}
\big]
\in\mathbb{R}^{d_\ell\times k},
\end{aligned}
\label{eq:signature}
\end{equation}
where $X_\ell^{(i)}\in\mathbb{R}^{d_\ell\times n_i}$ collects the $n_i$ token-level inputs to layer $\ell$ from episode $i$ as columns, and $(\sigma_1,u_1),\dots,(\sigma_k,u_k)$ are the top-$k$ eigenpairs of $G_\ell^{(t)}$.
Only $U_\ell^{(t)}$ is stored, giving each task a signature of $\sum_\ell k d_\ell$ values.
Because activations are accumulated into a single sum, no individual training sample is retained.
At the start of a deployment episode, the same sketch $q_\ell$ is computed from the initial observation and instruction, and the task is selected by
\begin{equation}
t^*
=
\arg\max_t
\frac{1}{L}
\sum_\ell
\mathrm{sim}\big(q_\ell,U_\ell^{(t)}\big).
\label{eq:sim}
\end{equation}
We use the uncentered variant of linear CKA~\cite{kornblith2019similarity},
\begin{equation}
\mathrm{sim}(q,U)
=
\frac{\|q^\top U\|_F^2}
{\|q^\top q\|_F\,\|U^\top U\|_F}.
\label{eq:cka}
\end{equation}
This compares the Gram matrices represented by the two sketches without explicitly storing them.
Centering is omitted because the accumulation in Eq.~\eqref{eq:signature} does not retain first-moment statistics.
Task selection requires one adapter-free forward pass followed by comparisons with the $T$ stored signatures, and is performed only once per episode.
The frozen backbone used to construct the signatures remains unchanged throughout the task stream, so signatures from earlier tasks remain valid.


\begin{table*}[t]
\vspace{2mm}
\centering
\caption{\textbf{Results on the four LIBERO suites} (3 seeds, mean\textsubscript{$\pm$sd}).
\method{} attains the highest $\mathrm{SR}_{\mathrm{final}}$ and AUC on all suites with competitive FWT.}
\label{tab:main}
\setlength{\tabcolsep}{2.5pt}
\newcommand{\ms}[2]{#1\,{\scriptsize$\pm$#2}}
\resizebox{\textwidth}{!}{%
\begin{tabular}{l cccc cccc cccc cccc}
\toprule
& \multicolumn{4}{c}{LIBERO-Long} & \multicolumn{4}{c}{LIBERO-Goal} & \multicolumn{4}{c}{LIBERO-Spatial} & \multicolumn{4}{c}{LIBERO-Object} \\
\cmidrule(lr){2-5} \cmidrule(lr){6-9} \cmidrule(lr){10-13} \cmidrule(lr){14-17}
Method & FWT$\uparrow$ & SR$_{\mathrm{final}}$$\uparrow$ & NBT$\downarrow$ & AUC$\uparrow$
       & FWT$\uparrow$ & SR$_{\mathrm{final}}$$\uparrow$ & NBT$\downarrow$ & AUC$\uparrow$
       & FWT$\uparrow$ & SR$_{\mathrm{final}}$$\uparrow$ & NBT$\downarrow$ & AUC$\uparrow$
       & FWT$\uparrow$ & SR$_{\mathrm{final}}$$\uparrow$ & NBT$\downarrow$ & AUC$\uparrow$ \\
\midrule
SeqFFT   & \ms{70.1}{2.0} & \ms{8.8}{0.2} & \ms{74.7}{0.8} & \ms{20.9}{0.0}
         & \best{\ms{90.7}{1.0}} & \ms{9.4}{0.5} & \ms{87.6}{1.4} & \ms{25.6}{0.5}
         & \best{\ms{87.0}{0.8}} & \ms{9.2}{0.9} & \ms{82.7}{0.7} & \ms{26.2}{0.5}
         & \best{\ms{93.2}{1.7}} & \ms{21.8}{1.1} & \ms{78.4}{0.3} & \ms{36.2}{0.6} \\
SeqLoRA  & \ms{57.0}{0.8} & \ms{4.1}{0.9} & \ms{66.5}{1.9} & \ms{15.7}{0.4}
         & \ms{82.4}{0.3} & \ms{6.8}{1.8} & \ms{83.7}{0.3} & \ms{23.4}{0.1}
         & \ms{75.6}{0.7} & \ms{7.8}{0.5} & \ms{77.9}{2.0} & \ms{22.4}{0.5}
         & \ms{82.9}{0.2} & \ms{15.0}{1.2} & \ms{75.9}{1.5} & \ms{31.1}{1.1} \\
PackNet  & \ms{12.4}{2.8} & \ms{0.0}{0.0} & \ms{26.3}{4.0} & \ms{1.5}{0.4}
         & \ms{25.9}{1.8} & \ms{0.0}{0.0} & \ms{40.8}{2.9} & \ms{3.8}{0.2}
         & \ms{23.5}{1.7} & \ms{0.0}{0.0} & \ms{35.7}{2.8} & \ms{2.8}{0.1}
         & \ms{15.2}{1.4} & \ms{0.0}{0.0} & \ms{26.3}{4.9} & \ms{1.7}{0.2} \\
EWC      & \ms{60.3}{0.4} & \ms{5.7}{0.8} & \ms{70.3}{1.5} & \ms{16.8}{0.9}
         & \ms{83.6}{1.2} & \ms{5.4}{1.2} & \ms{85.3}{0.7} & \ms{22.9}{1.0}
         & \ms{83.8}{0.9} & \ms{9.2}{1.4} & \ms{83.1}{0.6} & \ms{25.3}{0.2}
         & \ms{90.7}{0.7} & \ms{26.5}{4.0} & \ms{74.1}{1.4} & \ms{38.7}{0.6} \\
ER       & \ms{64.5}{3.4} & \ms{61.5}{3.8} & \ms{18.2}{1.6} & \ms{61.6}{3.5}
         & \ms{89.1}{0.3} & \ms{83.2}{1.5} & \ms{10.3}{2.1} & \ms{85.8}{1.6}
         & \ms{85.8}{0.3} & \ms{81.2}{1.8} & \ms{11.6}{2.0} & \ms{81.8}{1.7}
         & \ms{92.0}{0.9} & \ms{90.1}{2.5} & \ms{6.5}{1.7} & \ms{91.8}{1.4} \\
SHARE    & \ms{30.4}{1.7} & \ms{1.2}{0.4} & \ms{43.0}{3.7} & \ms{8.3}{0.4}
         & \ms{49.4}{4.1} & \ms{0.8}{1.2} & \ms{58.4}{8.5} & \ms{16.3}{0.9}
         & \ms{43.9}{3.1} & \ms{0.0}{0.0} & \ms{54.1}{4.0} & \ms{14.6}{0.7}
         & \ms{47.6}{2.8} & \ms{0.0}{0.0} & \ms{57.0}{1.5} & \ms{16.4}{2.4} \\
CLARE    & \ms{71.4}{1.6} & \ms{60.3}{8.3} & \ms{25.5}{8.3} & \ms{61.9}{4.9}
         & \ms{90.1}{1.0} & \ms{84.2}{6.7} & \ms{11.6}{5.5} & \ms{84.9}{5.3}
         & \ms{82.4}{0.3} & \ms{82.8}{0.8} & \ms{7.4}{0.9}  & \ms{83.0}{0.4}
         & \ms{88.4}{0.8} & \ms{87.8}{2.1} & \ms{9.0}{1.5}  & \ms{88.2}{0.8} \\
DMPEL    & \best{\ms{73.4}{0.6}} & \ms{72.9}{1.8} & \ms{11.1}{0.7} & \ms{74.8}{1.1}
         & \ms{88.3}{0.8} & \ms{72.8}{10.0} & \ms{8.7}{3.4}  & \ms{86.9}{3.2}
         & \ms{83.2}{1.5} & \ms{69.8}{2.0} & \best{\ms{6.5}{1.7}} & \ms{84.3}{0.7}
         & \ms{78.6}{5.4} & \ms{72.2}{6.4} & \ms{9.1}{1.4}  & \ms{80.6}{6.3} \\

\midrule
\method{} (ours)
         & \ms{73.0}{1.2} & \best{\ms{83.7}{1.6}} & \best{\ms{7.5}{0.9}} & \best{\ms{80.9}{1.7}}
         & \ms{86.5}{2.2} & \best{\ms{90.9}{2.0}} & \best{\ms{6.6}{0.9}} & \best{\ms{89.1}{2.1}}
         & \ms{83.8}{1.9} & \best{\ms{87.6}{3.1}} & \ms{6.8}{2.4} & \best{\ms{86.4}{3.1}}
         & \ms{87.2}{8.0} & \best{\ms{95.3}{1.9}} & \best{\ms{3.1}{1.4}} & \best{\ms{92.9}{2.4}} \\
\bottomrule
\end{tabular}}
\end{table*}

\begin{figure*}[t]
    \centering
    \includegraphics[width=\linewidth]{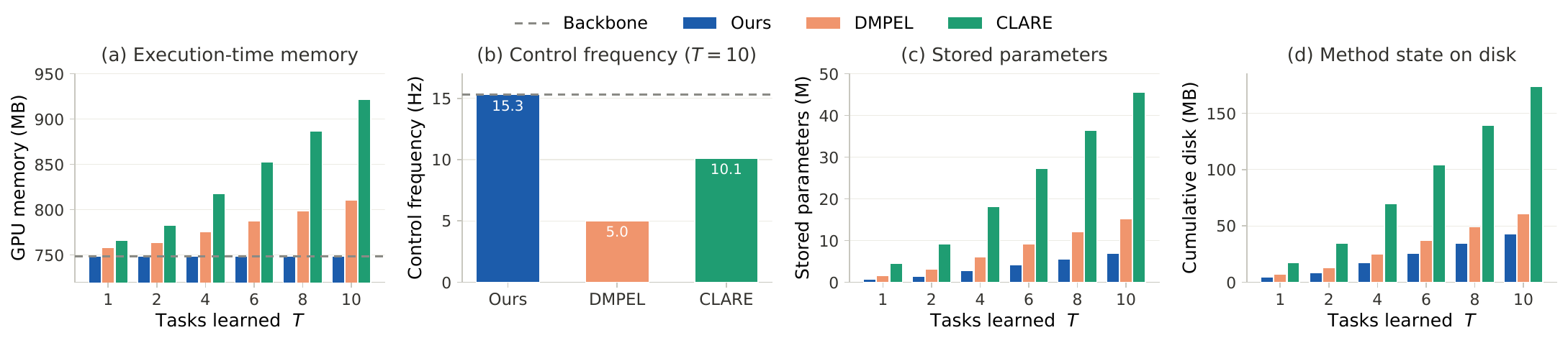}
    \caption{\textbf{Deployment cost versus tasks learned.}
    GPU memory and control frequency stay at the backbone's level for \method{}, and its stored parameters and on-disk state grow more slowly than adapter-based baselines.}
    \label{fig:deployment_cost}
\end{figure*}

\subsection{Deployment by Merging into the Backbone}
\label{sec:merging}
Given $t^{*}$, the update of Eq.~\eqref{eq:update} is evaluated from the stored combination coefficients and adapters and merged into the backbone,
\begin{equation}
W_\ell \;\leftarrow\; W_{0,\ell} + \Delta W^{(t^{*})}_\ell .
\label{eq:merge}
\end{equation}
The merge is exact because $\Delta W^{(t^{*})}_\ell$ is a single linear weight update, and it holds for the whole episode because $t^{*}$ is fixed at the start and $c^{(t^{*})}$ does not depend on the input.

After Eq.~\eqref{eq:merge}, no adapter, coefficient, or selection module is present in the control loop, and the resident model is a single backbone with modified weights.
Its memory footprint and per-step cost are therefore those of the backbone, regardless of $T$.
Merging requires an update that is linear in the weights and fixed for the episode.
Prior adapter-based methods~\cite{romer2026clare,lei2025dynamic} perform input-dependent adapter selection during execution, so their original inference does not define a single $\Delta W$ fixed for the entire episode.

\section{EXPERIMENTS}
\label{sec:exp}

Our experiments address three questions.
(i) Does \method{} retain previously learned tasks while acquiring new ones without the oracle task ID (Sec.~\ref{sec:main_results}, \ref{sec:real})?
(ii) Does it limit on-disk growth and keep GPU memory and control frequency at the backbone's level as tasks accumulate (Sec.~\ref{sec:deployment_cost})?
(iii) Which components of \method{} account for its performance (Sec.~\ref{sec:analysis})?

\subsection{Simulation Experimental Setup}
\label{sec:setup}
\noindent \textbf{Benchmark and protocol.} We evaluate on the four LIBERO suites~\cite{liu2023libero}: Long, Goal, Spatial, and Object, each a stream of $T{=}10$ tasks learned sequentially.
Following the LIBERO protocol, each task is probed with rollout episodes at regular intervals during training, and the best checkpoint is deployed as the task's final policy.
Every method runs three seeds under a shared schedule: 10k training steps per task, 20-episode probes every 800 steps, and 40 episodes for each retention evaluation. For \method{}, the 10k-step budget includes all three phases. Retention evaluations use each method's own task selection without the oracle task ID.

\noindent \textbf{Metrics.} Let $s_{j,t}$ be the success rate of task $\tau_t$ evaluated after the stream has learned $\tau_1$ through $\tau_j$.
Let $\bar s_t$ and $\hat s_t$ be the mean and best success rate of the rollout probes recorded while $\tau_t$ itself trains with known task identity; the best checkpoint is the one deployed.
We report the LIBERO metrics~\cite{liu2023libero}:
$\mathrm{FWT}=\frac{1}{T}\sum_{t\in[T]}\bar s_t$, how well each task is learned during training;
$\mathrm{NBT}=\frac{1}{T}\sum_{t=1}^{T}\mathrm{NBT}_t$, where
$\mathrm{NBT}_T=0$ and
$\mathrm{NBT}_t=\frac{1}{T-t}\sum_{j=t+1}^{T}(\hat s_t-s_{j,t})$
for $t<T$, the average drop from the deployed score when earlier tasks are revisited, so lower is better;
and $\mathrm{AUC}=\frac{1}{T}\sum_{t\in[T]}\frac{1}{T-t+1}\big(\bar s_t + \sum_{j>t} s_{j,t}\big)$, which summarizes the whole learn-then-retain trajectory.
We additionally report $\mathrm{SR}_{\mathrm{learn}}=\frac{1}{T}\sum_{t\in[T]}\hat s_t$, each task's success right after it is learned, and $\mathrm{SR}_{\mathrm{final}}=\frac{1}{T}\sum_{t\in[T]} s_{T,t}$, the average success over all tasks after the entire stream has been learned.

\noindent \textbf{Baselines.} We consider sequential adaptation baselines that overwrite previously learned parameters: SeqFFT, which fine-tunes the whole model for each new task, and SeqLoRA, which fine-tunes a single LoRA adapter across tasks.
We also consider EWC~\cite{kirkpatrick2017overcoming}, which regularizes parameters important to previous tasks, PackNet~\cite{mallya2018packnet}, which freezes the most important weights after each task and learns the next within the remaining free capacity, and ER~\cite{chaudhry2019tiny}, which retains past demonstrations and trains each new task on an even mix of old and new data.
We further include two adapter-based methods: CLARE~\cite{romer2026clare}, which trains task-specific adapters with per-task discriminators for selection, and DMPEL~\cite{lei2025dynamic}, which builds an expert library with a router fine-tuned by expert-coefficient replay.
SHARE~\cite{kaushik2026shared} represents resource-efficient continual learning methods that compress all tasks into a single adapter over a shared low-rank basis.
We include it to examine whether such methods, developed beyond robotics, transfer to lifelong robot learning.

\noindent \textbf{Implementation.}
All methods share one pre-trained VLA backbone: a DiT-based flow-matching policy~\cite{dasari2025ingredients} pretrained on LIBERO-90~\cite{liu2023libero}, with a DINOv2 vision encoder~\cite{oquab2023dinov2}, a CLIP text encoder~\cite{radford2021learning}, and an action head predicting 16-step chunks of which 8 are executed.
\method{} uses rank-8 LoRA adapters and one setting on every suite: sparsity-penalty weight of $\xi{=}2$ with an 800-step allocation phase and gates initialized from $\mathcal{N}(0,1)$ once per stream, duplication-penalty weight $\lambda{=}0.3$, and a coefficient pre-fit that stops when the loss plateaus.
Task signatures keep the top-2 eigenpairs per observation-side layer.
Our implementation builds on the official CLARE code~\cite{romer2026clare}, which provides the shared backbone and the SeqFFT, SeqLoRA, EWC, ER, and PackNet baselines.
DMPEL~\cite{lei2025dynamic} and SHARE~\cite{kaushik2026shared} are adapted from their official code with only the backbone replaced.
All baselines are reproduced under the protocol and metrics above.
All experiments run on a single NVIDIA A100 (80\,GB) GPU per task stream.

\begin{table}[t]
\vspace{2mm}
\centering
\caption{\textbf{Component ablations on LIBERO-Long}. 
$\Delta$ is AUC change from full.}
\label{tab:ablation}
\footnotesize
\setlength{\tabcolsep}{4pt}
\begin{tabular}{@{}l ccc r@{}}
\toprule
       & FWT$\uparrow$ & NBT$\downarrow$ & AUC$\uparrow$ & $\Delta$ \\
\midrule
\method{} (full) & 73.0$\pm$1.2 & 7.5$\pm$0.9 & 80.9$\pm$1.7 & \\
\midrule
w/o coefficient pre-fit & 68.7$\pm$1.7 & 6.8$\pm$0.9 & 79.0$\pm$0.5 & $-1.9$ \\
random allocation per task & 68.3$\pm$3.7 & 9.4$\pm$4.1 & 74.1$\pm$7.1 & $-6.7$ \\
fixed random allocation & 71.1$\pm$3.5 & 7.1$\pm$2.9 & 78.9$\pm$3.8 & $-2.0$ \\
fixed first-task allocation & 72.5$\pm$2.3 & 10.8$\pm$3.4 & 77.4$\pm$3.8 & $-3.5$ \\
w/o library reuse & 67.3$\pm$0.7 & 8.5$\pm$3.2 & 76.1$\pm$2.0 & $-4.7$ \\
w/o duplication penalty & 67.9$\pm$10.5 & 7.0$\pm$2.8 & 76.8$\pm$9.1 & $-4.1$ \\
\bottomrule
\end{tabular}
\end{table}

\begin{table}[t]
\centering
\caption{\textbf{Effect of the hyperparameters $\xi$ and $\lambda$ on LIBERO-Long}. Params denotes the cumulative number of stored parameters after ten tasks.
Larger $\xi$ trades performance for storage; the duplication penalty helps at moderate strength.}
\label{tab:hyperparameter}
\footnotesize
\setlength{\tabcolsep}{4.5pt}
\begin{tabular}{@{}cc ccc c@{}}
\toprule
$\xi$ & $\lambda$ & FWT$\uparrow$ & NBT$\downarrow$ & AUC$\uparrow$ & Params (M) \\
\midrule
2 & 0 & 67.9$\pm$10.5 & 7.0$\pm$2.8 & 76.8$\pm$9.1 & 6.8$\pm$2.0 \\
2 & 0.3 & 73.0$\pm$1.2 & 7.5$\pm$0.9 & 80.9$\pm$1.7 & 6.9$\pm$2.1 \\
2 & 0.6 & 68.1$\pm$8.1 & 10.0$\pm$2.7 & 75.3$\pm$4.1 & 6.8$\pm$1.9 \\
3 & 0.3 & 68.5$\pm$5.6 & 7.4$\pm$2.1 & 75.2$\pm$4.4 & 2.4$\pm$0.4 \\
4 & 0.3 & 59.8$\pm$5.0 & 11.6$\pm$3.5 & 64.3$\pm$4.2 & 1.0$\pm$0.2 \\
\bottomrule
\end{tabular}
\end{table}

\subsection{Main Results}
\label{sec:main_results}

Table~\ref{tab:main} shows the results on the four LIBERO suites.
\method{} achieves the highest $\mathrm{SR}_{\mathrm{final}}$ and AUC on all four suites and the lowest NBT on three of them, while maintaining competitive FWT.
Compared with adapter-based approaches, \method{} provides substantially stronger retention across the evaluated suites.
\method{} also outperforms ER in AUC across all suites without retaining past demonstrations.
SHARE performs poorly across all suites, suggesting that a single shared adapter is insufficient in this setting.
Overall, these results show that the proposed approach maintains both task acquisition and retention across the evaluated task streams.

\begin{figure}[t]
    \centering
    \includegraphics[width=\linewidth]{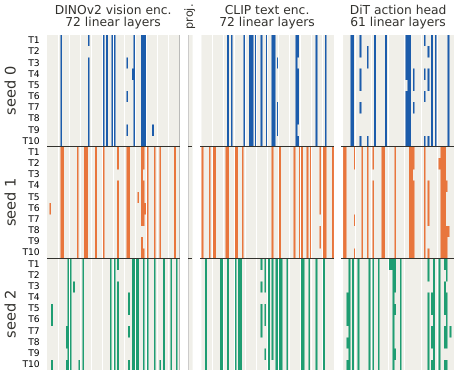}
    \caption{\textbf{Layers allocated on LIBERO-Long.}
    Colored bars mark layers that keep an adapter for each task (rows) and seed (blocks).}
    \label{fig:gate_survival}
\end{figure}

\subsection{Deployment Cost}
\label{sec:deployment_cost}
Figure~\ref{fig:deployment_cost} compares the deployment cost of the adapter-based methods as the task stream grows.
Because the selected update is merged before execution, \method{} runs at the backbone's own GPU memory and control frequency after ten tasks, whereas CLARE and DMPEL grow to 922 and 829\,MB and slow to 10.1 and 5.0\,Hz as their adapters and selection modules accumulate.
On disk, \method{} adds 3.8\,MB of task-specific state per task: 2.7\,MB for the newly allocated adapters, 1.1\,MB for the task signature, and the combination coefficients, one per layer per library adapter, which grow with the number of past tasks but remain below 10\,KB over the ten-task stream.
CLARE and DMPEL add 17.4 and 6.0\,MB per task.
Selection before the control loop adds little overhead.
Sketching the first observation, comparing it against the stored signatures, and merging the selected update take about 140\,ms in total and are paid once per episode.

\begin{figure*}[t]
    \centering
    \includegraphics[width=0.95\linewidth]{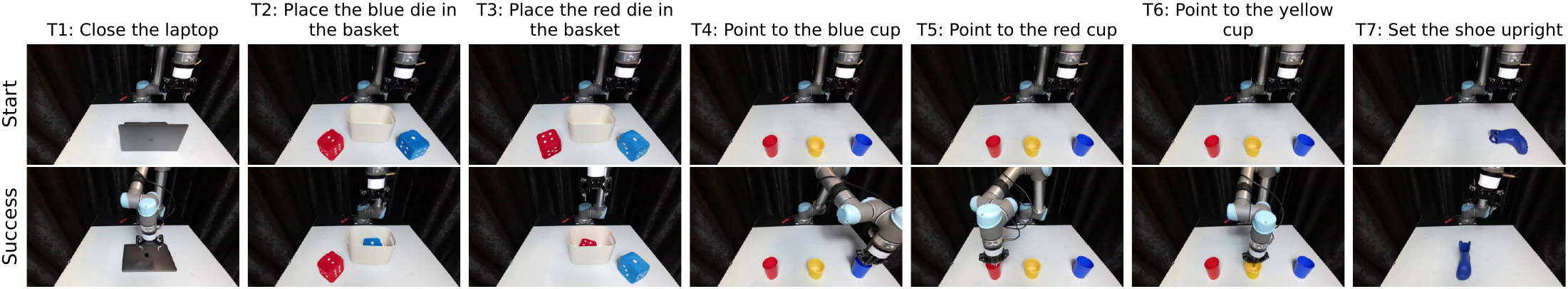}
    \caption{\textbf{Real-robot task stream.}
    Seven UR5 manipulation tasks learned sequentially.}
    \label{fig:real_tasks}
\end{figure*}

\subsection{Ablations and Analysis}
\label{sec:analysis}

\noindent \textbf{Component ablations.}
Table~\ref{tab:ablation} examines each component of \method{} on LIBERO-Long.
Without coefficient pre-fit, we skip Phase~1 and train the combination coefficients only with the new adapters in Phase~3.
This lowers FWT and AUC, suggesting that fitting reuse first helps isolate the residual for new adapters.
In \textit{random allocation per task}, we keep the same number of adapters as \method{} but randomly allocate their locations for each task.
This causes the largest AUC drop and substantially higher variation across seeds.
In \textit{fixed random allocation}, one randomly sampled layer set is shared by all tasks, recovering much of this loss and suggesting that shared adapter locations facilitate library reuse.
In \textit{fixed first-task allocation}, the gate-selected layers of the first task are reused throughout the stream.
Its lower AUC and higher variation than the full method indicate that task-specific gate fitting provides additional gains.
Without library reuse, we set $c^{(t)}{=}0$, preventing new tasks from using adapters learned for previous tasks.
The resulting drop in FWT and AUC shows that the frozen library provides useful capacity under the same sparse allocation budget.
Finally, without the duplication penalty, we set $\lambda{=}0$ during Phase~3.
This lowers mean FWT and AUC and substantially increases variation across seeds, indicating that the penalty improves both performance and stability.

\noindent \textbf{Where adapters are allocated.}
Figure~\ref{fig:gate_survival} shows the layers that survive gate-based allocation on LIBERO-Long for each task and seed.
Within each seed, tasks share many allocated layers under the common gate initialization, while task-specific gradients lead to differences at a few locations, and no two tasks end with an identical layer set.
Across seeds, the selected layers overlap little despite similar performance, suggesting that multiple sparse placements can be effective.
Together with Table~\ref{tab:ablation}, these results show that shared adapter locations facilitate library reuse, while fitting the gates for each task provides additional gains and stability.

\begin{figure}[t]
\vspace{2mm}
    \centering
    \includegraphics[width=0.95\linewidth]{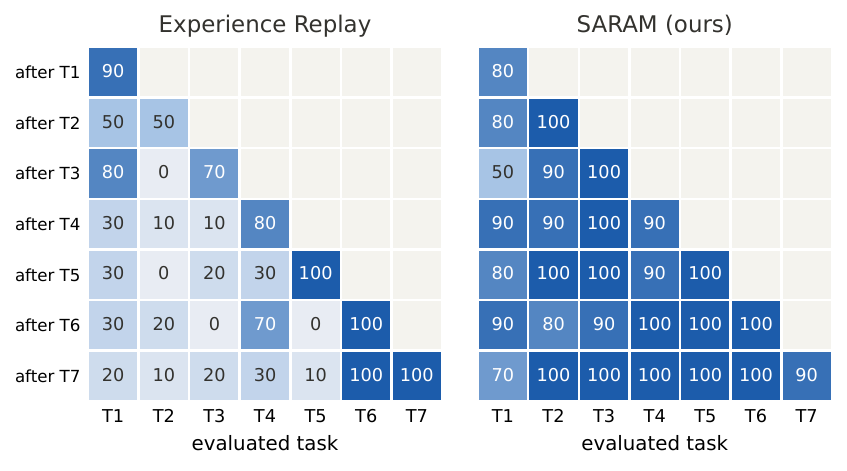}
    \caption{\textbf{Task performance across the UR5 stream.}}
    \label{fig:ur5_heatmap} 
\end{figure}

\begin{table}[t]
\centering
\caption{\textbf{Real-robot results on the UR5 stream}. 
$\mathrm{SR}_{\mathrm{learn}}$ is used as the NBT reference and the initial AUC point because no intermediate training probes are collected.}
\label{tab:real_robot_results}
\footnotesize
\setlength{\tabcolsep}{5pt}
\begin{tabular}{@{}l cccc@{}}
\toprule
Method & $\mathrm{SR}_{\mathrm{learn}}$$\uparrow$ & $\mathrm{SR}_{\mathrm{final}}$$\uparrow$ & NBT$\downarrow$ & AUC$\uparrow$ \\
\midrule
Per-task LoRA (oracle ref.) & 97.1 & -- & -- & -- \\
\midrule
SeqLoRA & 91.4 & 14.3 & 77.1 & 35.8 \\
ER & 84.3 & 41.4 & 40.2 & 53.6 \\
\method{} (ours) & \best{94.3} & \best{94.3} & \best{1.0} & \best{93.4} \\
\bottomrule
\end{tabular}
\end{table}

\begin{table}[t]
\centering
\caption{\textbf{Task-selection accuracy (\%) under instruction paraphrases.}
}
\label{tab:routing}
\footnotesize
\setlength{\tabcolsep}{4.5pt}
\begin{tabular}{l cccc c}
\toprule
& \multicolumn{4}{c}{Simulation (LIBERO)} & Real \\
\cmidrule(lr){2-5} \cmidrule(lr){6-6}
Instructions & Long & Goal & Spatial & Object & UR5 \\
\midrule
Original          & 100 & 100 & 100 & 100 & 100 \\
Mild paraphrase   & 100 & 100 & 100 & 100 & 100 \\
Strong paraphrase & 100 & 90  & 90  & 100 & 85.3 \\
\bottomrule
\end{tabular}
\end{table}

\noindent \textbf{Effect of $\xi$ and $\lambda$.}
Table~\ref{tab:hyperparameter} examines how the sparsity-penalty weight $\xi$ and duplication-penalty weight $\lambda$ affect performance and storage.
Increasing $\xi$ reduces the number of stored parameters at the cost of continual-learning performance, showing that it controls the trade-off between additional capacity and storage.
At a fixed sparsity level, the duplication strength $\lambda$ shows a clear performance trade-off.
A small $\lambda$ provides insufficient pressure to avoid duplicating reusable directions, whereas a large $\lambda$ can overly constrain the new adapters and reduce their ability to represent task-specific residuals.

\noindent \textbf{Task-signature robustness.}
Table~\ref{tab:routing} evaluates task recognition under unseen instruction paraphrases.
For this analysis only, we hold out 10 of the 50 episodes per LIBERO task (100 queries per suite) and the last 10\% of episodes per UR5 task (34 queries) from signature construction.
Signatures are built from the remaining episodes with the original instructions, and each held-out first observation is queried with either the original instruction or an unseen paraphrase.
A mild paraphrase changes only the action verbs, e.g., ``pick up the ketchup and place it in the basket'' becomes ``grab the ketchup and put it in the basket.''
A strong paraphrase restructures the sentence while preserving objects, attributes, and relations, e.g., ``point to the yellow cup'' becomes ``reach toward the cup that is yellow.''
Recognition remains accurate under mild paraphrases on all suites and the real-robot stream, and errors appear only under strong paraphrasing.
These results show that task recognition is not tied to the exact surface form of the instruction.

\subsection{Real-Robot Experiments}
\label{sec:real}
\noindent \textbf{Real-robot setup.}
We evaluate on a UR5 robot with a Robotiq 2F-85 gripper, an external camera, and a wrist camera.
The robot learns the seven tasks in Fig.~\ref{fig:real_tasks} sequentially.
Each task has about 50 teleoperated demonstrations and is evaluated with 10 rollouts.
We use $\pi_0$~\cite{black2024pi} as the backbone with rank-8 LoRA candidates on its linear layers and $\xi{=}40$, which keeps a similar fraction of layers per task as on LIBERO; other settings follow the simulation setup.
We compare against SeqLoRA, ER, and per-task LoRA policies selected with the oracle task ID as a reference.

\noindent \textbf{Real-robot results.}
Table~\ref{tab:real_robot_results} and Fig.~\ref{fig:ur5_heatmap} show the results over the seven-task stream.
SeqLoRA retains only the last task, and ER, despite replaying past demonstrations, loses most earlier tasks as the stream grows.
\method{} learns tasks close to the per-task reference and keeps them: its final success equals its success at learning time, with per-task differences within one episode.
Task selection is correct on every held-out episode, including the same-scene die and cup tasks (Table~\ref{tab:routing}).
These results match the LIBERO experiments and show that the framework transfers to a physical robot with a different VLA backbone.


\section{CONCLUSION}
\label{sec:conclusion}
We presented \method{}, a lifelong robot learning method for VLA policies that limits storage growth during learning and removes task-specific computation from the control loop.
Each new task reuses the frozen adapters of past tasks, adds new adapters only at layers selected by a sparsity penalty, and trains them under a duplication penalty.
At deployment, a task signature built from frozen-backbone activations selects the task once per episode without the oracle task ID or a learned selection module, and the selected update is merged into the backbone so that only the backbone runs in the control loop.
On four LIBERO suites, \method{} attains the highest $\mathrm{SR}_{\mathrm{final}}$ and AUC of all compared methods while adding up to $4.5\times$ less on-disk state per task than adapter-based baselines, and its GPU memory and control frequency stay at the backbone's level while the adapter-based baselines slow down by up to $3\times$.
On a seven-task real-robot stream with a different backbone, it learns and retains all seven tasks throughout the stream.





\bibliographystyle{IEEEtran}
\bibliography{ref}

@article{romer2026clare,
  title={Clare: Continual learning for vision-language-action models via autonomous adapter routing and expansion},
  author={R{\"o}mer, Ralf and Zhang, Yi and Li, Yuming and Schoellig, Angela P},
  journal={IEEE Robotics and Automation Letters},
  year={2026},
  publisher={IEEE}
}

@inproceedings{liu2024tail,
  title={Tail: Task-specific adapters for imitation learning with large pretrained models},
  author={Liu, Zuxin and Zhang, Jesse and Asadi, Kavosh and Liu, Yao and Zhao, Ding and Sabach, Shoham and Fakoor, Rasool},
  booktitle={International Conference on Learning Representations},
  volume={2024},
  pages={16330--16353},
  year={2024}
}

@article{lei2025dynamic,
  title={Dynamic mixture of progressive parameter-efficient expert library for lifelong robot learning},
  author={Lei, Yuheng and Mao, Sitong and Zhou, Shunbo and Zhang, Hongyuan and Li, Xuelong and Luo, Ping},
  journal={arXiv preprint arXiv:2506.05985},
  year={2025}
}

@article{liu2023libero,
  title={Libero: Benchmarking knowledge transfer for lifelong robot learning},
  author={Liu, Bo and Zhu, Yifeng and Gao, Chongkai and Feng, Yihao and Liu, Qiang and Zhu, Yuke and Stone, Peter},
  journal={Advances in Neural Information Processing Systems},
  volume={36},
  pages={44776--44791},
  year={2023}
}

@article{thrun1995lifelong,
  title={Lifelong robot learning},
  author={Thrun, Sebastian and Mitchell, Tom M},
  journal={Robotics and autonomous systems},
  volume={15},
  number={1-2},
  pages={25--46},
  year={1995},
  publisher={Elsevier}
}

@article{chaudhry2019tiny,
  title={On tiny episodic memories in continual learning},
  author={Chaudhry, Arslan and Rohrbach, Marcus and Elhoseiny, Mohamed and Ajanthan, Thalaiyasingam and Dokania, Puneet K and Torr, Philip HS and Ranzato, Marc'Aurelio},
  journal={arXiv preprint arXiv:1902.10486},
  year={2019}
}

@article{kirkpatrick2017overcoming,
  title={Overcoming catastrophic forgetting in neural networks},
  author={Kirkpatrick, James and Pascanu, Razvan and Rabinowitz, Neil and Veness, Joel and Desjardins, Guillaume and Rusu, Andrei A and Milan, Kieran and Quan, John and Ramalho, Tiago and Grabska-Barwinska, Agnieszka and others},
  journal={Proceedings of the national academy of sciences},
  volume={114},
  number={13},
  pages={3521--3526},
  year={2017},
  publisher={National Academy of Sciences}
}

@inproceedings{mallya2018packnet,
  title={Packnet: Adding multiple tasks to a single network by iterative pruning},
  author={Mallya, Arun and Lazebnik, Svetlana},
  booktitle={2018 IEEE/CVF Conference on Computer Vision and Pattern Recognition},
  pages={7765--7773},
  year={2018},
  organization={IEEE}
}

@article{hu2021lora,
  title={Lora: Low-rank adaptation of large language models},
  author={Hu, Edward J and Shen, Yelong and Wallis, Phillip and Allen-Zhu, Zeyuan and Li, Yuanzhi and Wang, Shean and Wang, Lu and Chen, Weizhu},
  journal={arXiv preprint arXiv:2106.09685},
  year={2021}
}

@inproceedings{kornblith2019similarity,
  title={Similarity of neural network representations revisited},
  author={Kornblith, Simon and Norouzi, Mohammad and Lee, Honglak and Hinton, Geoffrey},
  booktitle={International conference on machine learning},
  pages={3519--3529},
  year={2019},
  organization={PMlR}
}

@article{beck2009fast,
  title={A fast iterative shrinkage-thresholding algorithm for linear inverse problems},
  author={Beck, Amir and Teboulle, Marc},
  journal={SIAM journal on imaging sciences},
  volume={2},
  number={1},
  pages={183--202},
  year={2009},
  publisher={SIAM}
}

@inproceedings{dasari2025ingredients,
  title={The ingredients for robotic diffusion transformers},
  author={Dasari, Sudeep and Mees, Oier and Zhao, Sebastian and Srirama, Mohan Kumar and Levine, Sergey},
  booktitle={2025 IEEE International Conference on Robotics and Automation (ICRA)},
  pages={15617--15625},
  year={2025},
  organization={IEEE}
}

@article{oquab2023dinov2,
  title={Dinov2: Learning robust visual features without supervision},
  author={Oquab, Maxime and Darcet, Timoth{\'e}e and Moutakanni, Th{\'e}o and Vo, Huy and Szafraniec, Marc and Khalidov, Vasil and Fernandez, Pierre and Haziza, Daniel and Massa, Francisco and El-Nouby, Alaaeldin and others},
  journal={arXiv preprint arXiv:2304.07193},
  year={2023}
}

@inproceedings{radford2021learning,
  title={Learning transferable visual models from natural language supervision},
  author={Radford, Alec and Kim, Jong Wook and Hallacy, Chris and Ramesh, Aditya and Goh, Gabriel and Agarwal, Sandhini and Sastry, Girish and Askell, Amanda and Mishkin, Pamela and Clark, Jack and others},
  booktitle={International conference on machine learning},
  pages={8748--8763},
  year={2021},
  organization={PmLR}
}

@inproceedings{ding2023sparse,
  title={Sparse low-rank adaptation of pre-trained language models},
  author={Ding, Ning and Lv, Xingtai and Wang, Qiaosen and Chen, Yulin and Zhou, Bowen and Liu, Zhiyuan and Sun, Maosong},
  booktitle={Proceedings of the 2023 conference on empirical methods in natural language processing},
  pages={4133--4145},
  year={2023}
}

@article{kim2024openvla,
  title={Openvla: An open-source vision-language-action model},
  author={Kim, Moo Jin and Pertsch, Karl and Karamcheti, Siddharth and Xiao, Ted and Balakrishna, Ashwin and Nair, Suraj and Rafailov, Rafael and Foster, Ethan and Lam, Grace and Sanketi, Pannag and others},
  journal={arXiv preprint arXiv:2406.09246},
  year={2024}
}

@article{black2024pi,
  title={{$\pi_0$}: A Vision-Language-Action Flow Model for General Robot Control},
  author={Black, Kevin and Brown, Noah and Driess, Danny and Esmail, Adnan and Equi, Michael and Finn, Chelsea and Fusai, Niccolo and Groom, Lachy and Hausman, Karol and Ichter, Brian and others},
  journal={arXiv preprint arXiv:2410.24164},
  year={2024}
}

@article{intelligence2025pi,
  title={{$\pi_{0.5}$}: A Vision-Language-Action Model with Open-World Generalization},
  author={{Physical Intelligence} and Black, Kevin and Brown, Noah and Darpinian, James and Dhabalia, Karan and Driess, Danny and Esmail, Adnan and Equi, Michael and Finn, Chelsea and Fusai, Niccolo and others},
  journal={arXiv preprint arXiv:2504.16054},
  year={2025}
}

@article{saha2021gradient,
  title={Gradient projection memory for continual learning},
  author={Saha, Gobinda and Garg, Isha and Roy, Kaushik},
  journal={arXiv preprint arXiv:2103.09762},
  year={2021}
}

@article{kaushik2026shared,
  title={Shared LoRA Subspaces for almost Strict Continual Learning},
  author={Kaushik, Prakhar and Vaidya, Ankit and Chaudhari, Shravan and Chellappa, Rama and Yuille, Alan},
  journal={arXiv preprint arXiv:2602.06043},
  year={2026}
}

@inproceedings{qiao2026merge,
  title={Merge before forget: A single LoRA continual learning via continual merging},
  author={Qiao, Fuli and Mahdavi, Mehrdad},
  booktitle={International Conference on Learning Representations},
  volume={2026},
  pages={8133--8159},
  year={2026}
}

@inproceedings{fukuda2025adapter,
  title={Adapter merging with centroid prototype mapping for scalable class-incremental learning},
  author={Fukuda, Takuma and Kera, Hiroshi and Kawamoto, Kazuhiko},
  booktitle={2025 IEEE/CVF Conference on Computer Vision and Pattern Recognition (CVPR)},
  pages={4884--4893},
  year={2025},
  organization={IEEE}
}

@article{wang2024comprehensive,
  title={A comprehensive survey of continual learning: Theory, method and application},
  author={Wang, Liyuan and Zhang, Xingxing and Su, Hang and Zhu, Jun},
  journal={IEEE transactions on pattern analysis and machine intelligence},
  volume={46},
  number={8},
  pages={5362--5383},
  year={2024},
  publisher={IEEE}
}

@article{brohan2022rt,
  title={Rt-1: Robotics transformer for real-world control at scale},
  author={Brohan, Anthony and Brown, Noah and Carbajal, Justice and Chebotar, Yevgen and Dabis, Joseph and Finn, Chelsea and Gopalakrishnan, Keerthana and Hausman, Karol and Herzog, Alex and Hsu, Jasmine and others},
  journal={arXiv preprint arXiv:2212.06817},
  year={2022}
}

@article{brohan2023rt,
  title={Rt-2: Vision-language-action models transfer web knowledge to robotic control},
  author={Brohan, Anthony and Brown, Noah and Carbajal, Justice and Chebotar, Yevgen and Chen, Xi and Choromanski, Krzysztof and Ding, Tianli and Driess, Danny and Dubey, Avinava and Finn, Chelsea and others},
  journal={arXiv preprint arXiv:2307.15818},
  year={2023}
}

@article{goodfellow2013empirical,
  title={An empirical investigation of catastrophic forgetting in gradient-based neural networks},
  author={Goodfellow, Ian J and Mirza, Mehdi and Xiao, Da and Courville, Aaron and Bengio, Yoshua},
  journal={arXiv preprint arXiv:1312.6211},
  year={2013}
}

\end{document}